\documentclass[11pt,a4paper]{article}
\usepackage[hyperref]{acl2020}
\usepackage{times}
\usepackage{latexsym}

\usepackage{microtype}

\aclfinalcopy 

\usepackage{graphicx}
\usepackage{amsmath}
\usepackage{subcaption}
\usepackage{xcolor}
\usepackage{colortbl}
\usepackage{multirow}
\usepackage{cuted}
\usepackage[title]{appendix}

\definecolor{light-gray}{gray}{0.95}

\title{Behind the Mask: Demographic bias in name detection for PII masking}

\author{Courtney Mansfield$^*$, Amandalynne Paullada$^*\dagger$, Kristen Howell$^*$\\
  $^*$LivePerson Inc., Seattle, Washington, USA \\
  $^\dagger$Biomedical Informatics \& Medical Education, University of Washington, Seattle, Washington, USA\\
  \texttt{cmansfield@liveperson.com, paullada@uw.edu, khowell@liveperson.com}}

\date{}

\begin{document}
\maketitle


\begin{abstract}

Many datasets contain personally identifiable information, or PII, which poses privacy risks to individuals. PII masking is commonly used to redact personal information such as names, addresses, and phone numbers from text data. Most modern PII masking pipelines involve machine learning algorithms. However, these systems may vary in performance, such that individuals from particular demographic groups bear a higher risk for having their personal information exposed. In this paper, we evaluate the performance of three off-the-shelf PII masking systems on name detection and redaction. We generate data using names and templates from the customer service domain. We find that an open-source RoBERTa-based system shows fewer disparities than the commercial models we test. However, all systems demonstrate significant differences in error rate based on demographics. In particular, the highest error rates occurred for names associated with Black and Asian/Pacific Islander individuals.

\end{abstract}



\section{Introduction}

In a time of extensive data collection and distribution, privacy is a vitally important but elusive goal. In 2021, the US-based Identity Theft Resource Center reported a 68\% increase in data breaches from the previous year, with 83\% involving sensitive information{\footnote{\href{https://www.idtheftcenter.org/post/identity-theft-resource-center-2021-annual-data-breach-report-sets-new-record-for-number-of-compromises/}{https://www.idtheftcenter.org/post/identity-theft-resource-center-2021-annual-data-breach-report-sets-new-record-for-number-of-compromises/}}}. The exposure of personally identifiable information (PII), such as names, addresses, or social security numbers, leaves individuals vulnerable to identity theft and fraud. In response, a growing number of companies provide data protection services, including PII detection, redaction (masking), and anonymization.

PII masking offers assurances of security. However, this paper considers whether the models powering these services perform fairly across individuals, regardless of race, ethnicity, and gender. Historically, the US ``Right to Privacy'' concept has been centered around Whiteness, initially to protect White women from the then-emergent technology of photography and visual media \citep{osucha2009whiteness}. Black individuals have had less access to privacy and face greater risk of harm due to surveillance, including algorithmic surveillance \citep{browne2015dark, fagan2016stops}. 

In this paper, we evaluate the detection and masking of names, which are the primary indexer of a person's identity. We sample datasets of names and demographic information to measure the performance of off-the-shelf PII maskers. Although model bias or unfairness can be the result of a number of factors, including training data or presuppositions encoded in the algorithms themselves, the commercial systems we examine fail to provide details about training data or implementation. Therefore, we do not hypothesize a causal relationship between these factors and our findings.

Our work quantifies disparities in the name detection of PII masking systems where poor performance can directly and negatively impact individuals.  We demonstrate significant disparities in the recognition of names based on demographic characteristics, especially for names associated with Black and Asian/Pacific Islander groups.

\section{PII Masking}

This study analyzes personally identifiable information (PII) masking systems which aim to detect and redact sensitive personal information, particularly names, from text. This has been an important problem in the biomedical domain, in terms of preparing de-identified patient data for research \citep{kayaalp2018patient}, but is also increasingly important in an age of language models trained from web-scraped text, which have been shown to reveal private information that was not removed from the underlying training data \citep{carlini2021extracting}.

Since early efforts masking data by hand, automated methods have been employed, from using word lists or dictionaries \citep{thomas2002successful}, which do not generalize to unseen names and locations, to rule-based or regular expression systems \citep{beckwith2006development,friedlin2008software}, which are generalizable, but can be brittle. These have been replaced with machine learning systems \citep{szarvas2006multilingual,uzuner2008identifier} and most recently neural networks \citep{dernoncourt2017identification,adams2019anonymate}.

Modern PII maskers rely on Named Entity Recognition (NER) to identify entities (e.g. name and location) for redaction. NER has had recent success with hybrid bi-directional long short term memory (BiLSTM) and conditional random field (CRF) models \citep{huang2015bidirectional}, and following the general trend in NLP, fine-tuning on large language models such as BERT \citep{li2019unified}. Additional discussion on NER architectures can be found in \citet{li2020survey}.

Previous research in Named Entity Recognition (NER) has illuminated race and gender-based disparities. \citet{mishra2020assessing} evaluates a number of NER models which consider performance according to gender and race/ethnicity. The analysis considers 15 names per intersectional group, finding that White-associated names are more likely to be recognized across all systems. Our work differs from and extends this work in key aspects: focusing on off-the-shelf PII masking, providing analysis on over 4K names, and reporting on significance and additional metrics.

Recent PII masking models perform extremely well in certain contexts. The recurrent neural network of \citet{dernoncourt2017identification} achieves 99\% recall overall and just below 98\% for names on patient discharge summaries in the medical domain. The commercial models we consider do not advertise performance metrics, and as shown in Section~\ref{sec:results}, do not achieve such high performance across our datasets.

It is important to note that removing names alone is insufficient to fully protect individuals from being identified from data. Data sets can still reveal just enough information to re-identify individuals, as in the case of Massachusetts Governor William Weld, whose medical records, although not connected directly to his name in a de-identified data set, were traceable back to him by matching information from an easily attained external data resource \cite{sweeney2002k}. Here we focus on names as they are a primary identifier for an individual.

\section{What's in a Name?} 


The primary goal of this paper is to understand whether, and to what degree, the performance of PII masking models is influenced by correlates of race, ethnicity, and gender. We frame bias in terms of significant discrepancies in performance based on race/ethnicity and gender, looking specifically to instances where private information was not masked (false negative rates, described in Section \ref{sec:metrics}). PII masking is a primary mechanism for protecting personal data, and a systematic failure to mask information belonging to marginalized subgroups can cause undue harm to those populations, through identity theft, identity fraud, and loss of privacy. Names are not a proxy for gender or race/ethnicity, but our rationale is as follows: if most of the people with Name $N$ have self-identified as belonging to Group $G_1$, and Name $N$ is frequently miscategorized by PII systems at a rate that is higher than that for a name more commonly used by individuals in Group $G_2$, then we argue that members of Group $G_1$ bear a higher privacy risk. 

We focus our analysis on given names (sometimes known as `first names') and family names (sometimes known as `surnames' or `last names'). Naming conventions vary in different cultural and linguistic contexts. In many cultures, given names and/or family names can be gendered, or disproportionately associated with a particular gender, religious or ethnic group. In the present study, gender, race and ethnicity are considered with respect to a defined set of categories for the purpose of analysis, but we acknowledge that such labels are socially constructed and mutable over time and space \citep{sen2016race}.


Previous research has uncovered racial and gender discrimination  based on individual names. \citet{bertrand2004emily} found that, given identical resumes with only a change in name, resumes with Black-associated names received fewer callbacks than White-associated names. \citet{sweeney2013discrimination} found that internet searches for Black (in contrast to White) names were more likely to trigger advertisements that suggested the existence of arrest records for people with those names.


We do not attempt to infer personal information tied to names in our data, but rather, rely on real, self-reported information. However, there are limitations to using standardized gender and racial categories in studying algorithmic fairness, even when individuals are able to self-identify \citep{hanna2020towards}. Within each racial/ethnicity category made available on the standardized forms in the data we use (described in Section \ref{sec:data}), for example, there is a large variety in the linguistic cultures and naming practices encompassed in each group. Our intent is not to conflate race and ethnicity and language, but rather to get a coarse-grained look at performance of PII masking systems on names that are strongly associated with the demographic groupings that are available. Similarly, the available data limits gender categories to the binary `male' and `female,' and while names are not a good proxy for gender, we look for strong associations in the data, as described further in Section \ref{sec:data}.

\section{Data}\label{sec:data}

In this section, we describe our method for creating test sentences for evaluating name detection in PII masking models. In our evaluation, we use a sentence perturbation technique which is employed in previous studies to test model performance across sensitive groups \citep{garg2019counterfactual, hutchinson2020social}.  Using a variety of templates, we fill slots with names from the datasets, allowing us to measure performance across race/ethnicity and gender.

Reliable sources of demographically labeled names are difficult to find and using real names is an issue of privacy. Therefore, we consider datasets of names with aggregate demographic information as a proxy. We also evaluate on the names of US Congress members, whose identity and self-reported demographic information is publicly available. Templates and source datasets are described in the following sections.

\subsection{Templates}

We collected a set of 32 templates from real-world customer service messaging conversations (see examples in Table \ref{tab:template} and the full set in Appendix \ref{sec:appendix_templates}). These include dialog between customers and conversational AI or human agents. Customer service data is especially vulnerable to security threat, carrying potentially sensitive personal information such as credit card or social security numbers. Topics of discussion in the dataset include placing or tracking a purchase or paying a bill. Each template contains a name, which we replace with a generic \textsc{name} slot. Various identifiers from the dataset (e.g. location or reference numbers) 
are swapped to protect personal information. 

\begin{table}
\begin{center}
    \begin{tabular}{ |l| } 
         \hline
         Sample Templates \\ \hline
         This was from $<$\textsc{name}$>$ \\ \hline 
         The response is signed $<$\textsc{name}$>$ \\ \hline
         it’s YGDFEA the reservation. \\ $<$\textsc{name}$>$ \\ 
         \hline
    \end{tabular}
\caption{Sample of templates used for analysis.}
\label{tab:template}
\end{center}
\end{table}

\subsection{LAR Data}
The LAR dataset from \citet{tzioumis2018demographic} contains aggregate names with self-reported race/ethnicity from US Loan Application Registrars (LARs). It includes 4.2K given names from 2.6M observations across the US. Race/ethnicity categories are shown in Table \ref{tab:source-data-groups}.

There are limitations to the \citet{tzioumis2018demographic} dataset. Because the sample is drawn from mortgage applications and there are known racial and socioeconomic differences in who applies for mortgage applications \citep{charles2002transition}, the data is likely to contain representation bias. However, the LAR dataset is the largest available set of names and demographics, estimated to reflect 85.6\% of names in the US population \citep{tzioumis2018demographic}. Due to its large size, we are able to control for the frequency of names, as described in Section \ref{sec:sampling}.


\subsection{NYC Data}

The NYC dataset was created using the New York City (NYC) Department of Health and Mental Hygiene's civil birth registration data \citep{NYCOpenData} and contains 1.8K given names from 1.2M observations. Data is available from 2011-2018 and includes self-reported race/ethnicity of the birth mother (other parents' information is not available). The sex of the baby is included, which permits an intersectional analysis.\footnote{Although the NYC data includes the child's \textit{sex assigned at birth}, we use this variable to approximate the \textit{gender} associated with the name.} The race/ethnicity groups are shown in Table \ref{tab:source-data-groups}.

While the other datasets report on adult names, the NYC data aggregates the names of children who are between 4-11 at the time of this writing. This adds diversity in terms of age, as data privacy is an important issue for both children and adults.

\subsection{Congress Data}\label{subsec:congress}

The Congress dataset allows for evaluation over the given and family names of real individuals. The 540 current members of US Congress provide self-reported demographic information.\footnote{See \href{www.senate.gov}{www.senate.gov} and \\ \href{https://pressgallery.house.gov/member-data/demographics}{https://pressgallery.house.gov/member-data/demographics}.} 
Race/ethnic groups are described in Table \ref{tab:source-data-groups}. 76\% of congress members do not report membership in the race/ethnicity groups listed, and are grouped as ``White/Other''.

This dataset provides a naturalistic analysis of full names. Alternatively, one could programmatically generate given and family name pairs from datasets of first names and a dataset of last names. However, the broad race/ethnic groups used for classification do not account for the variance in the cultural backgrounds of the names (e.g. Pakistani and Native Hawaiian backgrounds are listed under the umbrella of Asian and Pacific Islander). 


\begin{table*}
\begin{center}
    \begin{tabular}{ |l|l|l| } 
         \hline
         Data & Dataset Race/Ethnicity Group & Mapped label \\ \hline
         LAR & NH Asian or Native Hawaiian or Other Pacific Islander & Asian and Pacific Islander \\
         & NH Black or African American & Black \\
         & Hispanic or Latino & Hispanic \\
         & NH American Indian or Alaska Native & Indigenous \\
         & NH Multi-race & Multi-race \\ 
         & NH White & White \\ \hline 
         NYC &  Asian and Pacific Islander & Asian and Pacific Islander \\
         & Black & Black \\
         & Hispanic White & Hispanic \\
         & NH White & White \\ \hline
         Cong. & Asian & Asian and Pacific Islander \\
         & Black & Black \\
         & Hispanic & Hispanic \\
         & Indigenous & Indigenous \\
         & White/Other & White \\ 
         \hline
    \end{tabular}
\caption{Race/ethnicity categories used for each data source and the mapped set of race/ethnic group labels each category is mapped to for our analysis. The term ``Non-Hispanic'' is abbreviated NH.}
\label{tab:source-data-groups}
\end{center}
\end{table*}

\section{Sampling Process} \label{sec:sampling}
This section describes the process of sampling the source names. The LAR and NYC datasets aggregate name counts and frequencies per race/ethnicity. We sample names which have a strong `association' with a particular race/ethnicity and gender. Because frequency (i.e. popularity) of a name could contribute to spurious performance disparities between groups, we sample the LAR data so that all names are frequency matched across groups.


\subsection{Demographic categorization}
For each group, we sample names that are ``associated'' with that particular group. We define ``association'' as when $>$75\% of people with the same name self-report within the same race/ethnicity. In the LAR dataset, the NH American Indian or Alaska Native and NH Multi-race names reflect 1\% of individuals in the dataset \citep{tzioumis2018demographic}. No names were found with strong associations in these groups, and for this reason, we do not include them in the analysis. We map race/ethnicity groups across datasets to a common set of labels, which are based on categories of the 2010 US Census dataset of surname and race/ethnicity information \citep{comenetz2016frequently}.  Race/ethnicity categorization for all datasets is shown in Table \ref{tab:source-data-groups}.

The NYC dataset also includes gender. Using a 90\% threshold for our definition of `association', 99\% of names in the source set are strongly association with one gender.

\subsection{Frequency matching}

Because the LAR dataset has a large sample size, it is possible to control for the frequency of names while maintaining a minimum threshold of 20 names per category. To standardize based on frequency, we use counts from the 2010 US Census Bureau. We did not use observation counts directly from the LAR data, due to the aforementioned potential for representational bias.

We sample the LAR dataset to align the mean observation counts of Black-associated names and other groups, as there are few Black-associated names in the dataset (n=21).  However, there is limited overlap in the frequency distributions of API-associated names with Hispanic and Black-associated names.  Therefore, we sample a second set with API and White-associated names only.  We refer to these datasets as LAR1 (Black, Hispanic, and White) and LAR2 (API and White).  The frequency matching process is described in more detail in Appendix \ref{sec:appendix_freq}.

\section{Experiment Setup}

The following sections discuss the PII masking systems we evaluate. We use several metrics to investigate the PII masking performance across name subsets.\footnote{Experiment code is publically available at \href{https://github.com/csmansfield/pii-masking-bias}{https://github.com/csmansfield/pii-masking-bias}.}

\subsection{Models}\label{sec:models}

We select two commercial and one open-source PII masking system for evaluation. The commercial systems we consider are Amazon Web Services (AWS) Comprehend and Google Cloud Platform Data Loss Prevention (GCP DLP). We choose these systems for their potentially large reach, with AWS and GCP holding a combined 43\% market share of cloud services.\footnote{\href{https://www.statista.com/chart/18819/worldwide-market-share-of-leading-cloud-infrastructure-service-providers/}{https://www.statista.com/chart/18819/worldwide-market-share-of-leading-cloud-infrastructure-service-providers/}} Amazon Comprehend provides an English model with a \textsc{Name} entity for PII redaction. GCP DLP offers redaction and includes a global \textsc{PERSON\_NAME} entity. Microsoft's Presidio is an open-source service for PII detection. We use the default English model which uses logic such as regex matching and Named Entity Recognition (NER). For the Presidio model we use a spaCy 3.2 en\_core\_web\_trf model for NER, which utilizes the RoBERTa-base Transformer model trained on OntoNotes 5.


\subsection{Evaluation metrics}\label{sec:metrics}


We measure false negative rates (FNRs), the rate at which a PII system does not detect a name that is present in the dataset (and therefore is unable to mask it).\footnote{
Whereas false positive rates are useful for evaluating the precision of a model, our focus is the failure to detect person names, rather than the incorrect identification of tokens that are not person names. Furthermore, we report no false positives in our findings.}  Following \citet{dixon2018measuring} we report on the False Negative Equality Difference, which measures differences between the false negative rate over the entire dataset and across each demographic subgroup $g$. We add a normalization term to compare the FNED of datasets with different numbers of groups, as shown in equation \ref{eq:normFNER}.

\begin{equation}
    \frac{1}{|G|}\sum_{g\in G} |FNR-FNR_{g}|
    \label{eq:normFNER}
\end{equation}

We also measure the statistical significance of performance differences across subgroups. We conduct Friedman and Wilcoxon signed-rank tests following \citet{czarnowska2021quantifying}.  The Friedman test is used for cases with more than 2 subgroups, and provides a single \textit{p}-value for each dataset and system pair. The \textit{p}-value determines whether to reject the null hypothesis that FNR of a given system is the same across all demographic groups. The statistic is calculated considering $j$ demographic subsets $g$. First, we calculate the average FNR for a template $t$, over all names belonging to a particular subset $g$. The averages for each of the 32 templates considering group $g$ are contained in $X_{g}$. The Friedman statistic is calculated for all $X_{g}$.

\begin{align}
    & X_{g} = (FNR(x_{g}^{1}),...,FNR(x_{g}^{32})) \nonumber \\
    & Friedman(X_{1},..., X_{j})
    \label{eq:sig}
\end{align}

\noindent Nemenyi post-hoc testing is used for further pairwise analysis. For cases with only 2 subgroups, we alternatively perform Wilcoxon signed-rank tests. In order to control for multiple comparisons, we apply a Bonferroni correction across all \textit{p}-values (at p$<$0.05 and n=15, our adjusted significance threshold is 0.003).

\section{Results}\label{sec:results}

We present the results of the evaluation, considering overall performance and performance related to race/ethnicity, gender, and intersectional factors.  The section concludes with an analysis of errors.

\subsection{Overall Performance}

The average performance on the datasets can be seen in Table \ref{tab:means_race}. System performance varies according to the dataset, with no single system performing best on all sets. All systems have lower FNR on the Congress dataset, where both given and family names are available, likely due to the increased information load of full names. The LAR2 and NYC names prove the most challenging across all systems.

\begin{table}
\begin{center}
   \begin{tabular}{ |l|l|r|r|r|r| } 
         \hline
          & Group & N &  \multicolumn{3}{c|}{FNR (\%)} \\ 
          & & & \multicolumn{1}{c}{AWS} & \multicolumn{1}{c}{GCP} & \multicolumn{1}{c|}{MP} \\ \hline
         LAR1 & Black & 20 & 20.0 & 18.1 & \textbf{29.5} \\
          & Hisp. & 172 & \textbf{28.4} & 12.4 & 24.7 \\
          & White & 1000 & 21.3 & \textbf{18.5} & 20.0 \\
          & \cellcolor{light-gray}All & \cellcolor{light-gray}1192 & \cellcolor{light-gray}22.3 & \cellcolor{light-gray}17.6 & \cellcolor{light-gray}20.8 \\
          \hline
         LAR2 & API & 441 & \textbf{38.2} & \textbf{51.2} & \textbf{29.2} \\
          & White & 1000 & 25.3 & 18.6 & 25.8 \\ 
          & \cellcolor{light-gray}All & \cellcolor{light-gray}1441 & \cellcolor{light-gray}29.3 & \cellcolor{light-gray}28.6 & \cellcolor{light-gray}26.8 \\
          \hline
          NYC & API & 165 & 21.3 & 43.6 & 22.0 \\
           & Black & 226 & \textbf{28.9} & \textbf{56.3} & \textbf{32.6} \\
           & Hisp. & 389 & 20.1 & 34.2 & 21.2 \\
           & White & 592 & 26.9 & 29.2 & 25.9 \\
          & \cellcolor{light-gray}All & \cellcolor{light-gray}1359 & \cellcolor{light-gray}24.6 & \cellcolor{light-gray}36.8 & \cellcolor{light-gray}25.2 \\
          \hline 
          Cong. & API & 16 & \textbf{23.0} & \textbf{12.1} & \textbf{11.7} \\
           & Black & 56 & 15.2 & 9.7 & 9.5 \\
           & Hisp. & 48 & 13.9 & 8.3 & 9.4 \\
           & Indig.$\dagger$ & 3 & 7.0 & 6.3 & 7.8 \\
           & Multi.$\dagger$ & 6 & 8.3 & 6.3 & 10.9 \\
           & White/ & 419 & 12.1 & 6.7 & 7.7 \\
           & Other & & & & \\
          & \cellcolor{light-gray}All & \cellcolor{light-gray}530 & \cellcolor{light-gray}12.8 & \cellcolor{light-gray}7.3 & \cellcolor{light-gray}8.1 \\
           \hline
    \end{tabular}
\caption{Support and average false negative rate (FNR) by race/ethnicity group across datasets. Groups marked with `$\dagger$' are not included in formal statistical analysis due to low support.  Maximum FNR per dataset/sytem is shown in bold.}
\label{tab:means_race}
\end{center}
\end{table}

\begin{figure}[ht]
    \includegraphics[width=0.48\textwidth]{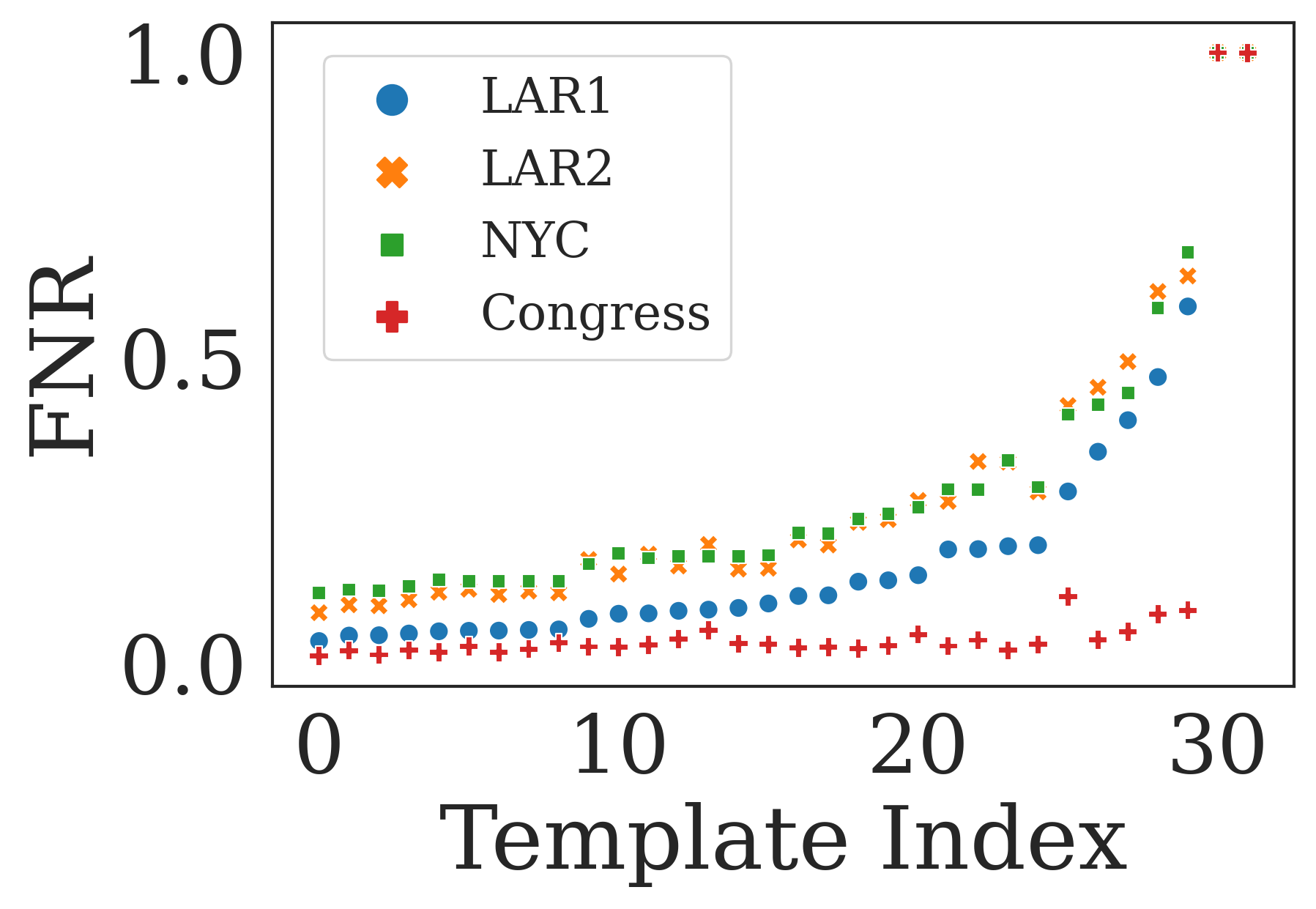}
    \caption{Average FNR across each template per dataset.}
    \label{fig:template_performance}
\end{figure}

The average performance of the names per each template is shown in Figure \ref{fig:template_performance}.  Performance varies considerably, with average FNR per template ranging between 6\%. and 100\%. The mean FNR for all templates is 22\%.

\subsection{Performance by Race/Ethnicity}

The normalized false negative equality differences (FNEDs) are shown in Table \ref{tab:FNED}.  

\begin{table}
\begin{center}
    \begin{tabular}{ |ll|r|r|r| }
         \hline
         & & \multicolumn{3}{c|}{FNED} \\
         & & \multicolumn{1}{c}{AWS} & \multicolumn{1}{c}{GCP} & \multicolumn{1}{c|}{MP} \\ \hline
         & LAR1 & *3.1 & *2.2 & \textbf{*4.4} \\
         Race/ & LAR2 & \textbf{*6.4} & \textbf{*16.3} & 1.7 \\ 
         ethnicity & NYC & *3.6 & *8.9 & *3.7 \\ 
         & Congress & *3.6 & *2.2 & 1.7 \\ \hline
        \multirow{ 2}{*}{Gender} & NYC & *3.2 & *4.4 & 0.8 \\ 
         & Congress & *1.3 & *0.6 & 0.2 \\ 
         \hline
    \end{tabular}
\caption{The normalized false negative equality difference (FNED) for race/ethnicity and gender subsets of the data. Asterisks indicate significance (p$<$0.003) in FNR differences by group. Maximum FNED per system is shown in bold.}
\label{tab:FNED}
\end{center}
\end{table}

The highest FNED, which is an 82\% increase over the second highest FNED, is seen in GCP's performance over the LAR2 dataset which includes frequency controlled API and White-associated names. The FNRs in Table \ref{tab:means_race} show high FNR for API names in LAR2 across all systems. The error rate for GCP is 175\% higher for API-associated names in this set. A Wilcoxon signed-rank test shows significant differences in FNR for AWS and GCP, with better performance on White-associated names. The Presidio transformer model has a smaller gap which is not found to be significant.

Performance on LAR1, which includes frequency-balanced Black, Hispanic, and White-associated names, also shows variability in FNR across race/ethnicity groups. However, the performance differences across groups are dependent on the system. For example, the Presidio transformer model shows poor performance on Black-associated names, and post-hoc tests (see Appendix \ref{sec:posthoc}) reveal significant differences between Black vs. Hispanic and White groups. On the other hand, AWS performs best on Black-associated names but significantly worse on Hispanic-associated names. GCP peforms worst on White-associated names.

The NYC dataset shows more consistency in terms of performance across groups, with Black-associated names having higher FNRs across all systems. This is further confirmed by statistical testing on AWS and GCP, where Black-associated names have statistically higher FNR than Hispanic-associated names. GCP also performs significantly worse on Black-associated names than White-associated names. Although significant FNR differences are found in the performance of Presidio on the basis of race/ethnicity, post-hoc tests did not indicate pair(s) which met the threshold for significance.

Finally, the Congress dataset, which includes given and family names, has the lowest FNED rates in terms of race/ethnicity. However, there are still significant differences in performance across groups for AWS and GCP maskers. Here, API-associated names again show high FNRs. Friedman tests and post-hoc testing support differences between API and other groups in the case of AWS and GCP. Performance on Black-associated names was also significantly worse than on White-associated names for GCP. There were no significant differences associated with the Presidio model.

\subsection{Performance by Gender}

The NYC and Congress datasets also include information about gender, which allows for a comparison of gender-based subsets. The FNEDs in Table \ref{tab:FNED} are generally lower for gender than for race. However, some gender-based differences are shown to be significant. 

The average FNR grouped by gender is shown in Table \ref{tab:means_gender}. The NYC dataset shows female-associated, male-associated, and `other' names, which are not strongly associated with a particular gender. FNR is highest for such unassociated names. Performance on female and male-associated names varies, with AWS performing significantly better on female-associated names, and GCP performing significantly better on male-associated names.

\begin{table}
\begin{center}
    \begin{tabular}{ |l|l|r|r|r|r| } 
         \hline
          & Gender & N & \multicolumn{3}{c|}{FNR (\%)} \\
          & & & \multicolumn{1}{c}{AWS} & \multicolumn{1}{c}{GCP} & \multicolumn{1}{c|}{MP} \\
         \hline
          NYC & F & 741 & 23.7 & 39.8 & 25.1 \\
           & M & 618 & 25.6 & 33.1 & 25.3 \\ 
           & Other $\dagger$ & 13 & \textbf{32.2} & \textbf{43.3} & \textbf{27.4} \\
           & \cellcolor{light-gray}All & \cellcolor{light-gray}1359 & \cellcolor{light-gray}24.5 & \cellcolor{light-gray}36.8 & \cellcolor{light-gray}25.2 \\ \hline
          Cong. & F & 145 & 11.0 & \textbf{8.2} & \textbf{10.0} \\
           & M & 385 & \textbf{13.6} & 7.0 & 8.5 \\
          & \cellcolor{light-gray}All & \cellcolor{light-gray}530 & \cellcolor{light-gray}12.9 & \cellcolor{light-gray}7.3 & \cellcolor{light-gray}8.9 \\
           \hline
    \end{tabular}
\caption{Support and average false negative rate (FNR) by gender across datasets. `Other' specifies names which are not strongly associated with one gender.  Groups marked with `$\dagger$' are not included in formal statistical analysis due to low support. Maximum FNR per dataset/system is shown in bold.}
\label{tab:means_gender}
\end{center}
\end{table}

\subsection{Intersectional Analysis} \label{sec:intersectional_results}

We analyzed the NYC results for differences across both race/ethnicity and gender. Table \ref{tab:means_nyc} shows FNR averages associated with intersectional groups. FNR for Black female-associated names is highest among all groups, and error rates are on average 13.7\% higher than that of the full dataset. Black male-associated names have the second highest FNR for GCP and MP. Pairwise testing does not reveal significant differences between Black male and female-associated names. The subsets with the lowest FNR vary across systems. Hispanic-associated names have the lowest FNR in AWS and Presidio. For GCP, White male-associated names have the lowest FNR. 

\begin{table}
\begin{center}
    \begin{tabular}{ |l|l|r|r|r|r| } 
         \hline
          Group & Gender & N &  \multicolumn{3}{c|}{FNR (\%)} \\
          & & & \multicolumn{1}{c}{AWS} & \multicolumn{1}{c}{GCP} & \multicolumn{1}{c|}{MP} \\
         \hline
         API & F & 86 & 20.1 & 43.0 & 22.2  \\
          & M & 77 & 22.1 & 43.9 & 22.2 \\ 
         Black & F & 122 & \textbf{30.1} & \textbf{62.8} & \textbf{34.7} \\
          & M & 101 & 27.0 & 47.2 & 29.2 \\ 
         Hisp. & F & 212 & 18.4 & 35.7 & 21.3 \\
          & M & 175 & 22.2 & 32.2 & 21.1 \\ 
         White & F & 321 & 25.7 & 32.9 & 24.8 \\
          & M & 265 & 28.2 & 25.2 & 27.4 \\
         \cellcolor{light-gray}All & \cellcolor{light-gray}- & \cellcolor{light-gray}1359 & \cellcolor{light-gray}24.5 & \cellcolor{light-gray}36.8 & \cellcolor{light-gray}25.2 \\
         \hline 
    \end{tabular}
\caption{Support and average false negative rate (FNR) by race/ethnicity and gender in the NYC dataset. Maximum FNR per system is shown in bold.}
\label{tab:means_nyc}
\end{center}
\end{table}

\subsection{Analysis of Names}

The previous findings in this section captured a few general patterns. One pattern that held across most systems and datasets was high false negative rates of API names. In the LAR2 and Congressional datasets, API names were especially hard for systems to detect. This was not simply due to API names being less common, as the LAR2 set included names balanced by their frequency in the general US population.

\begin{figure}[ht]
    \includegraphics[width=0.48\textwidth]{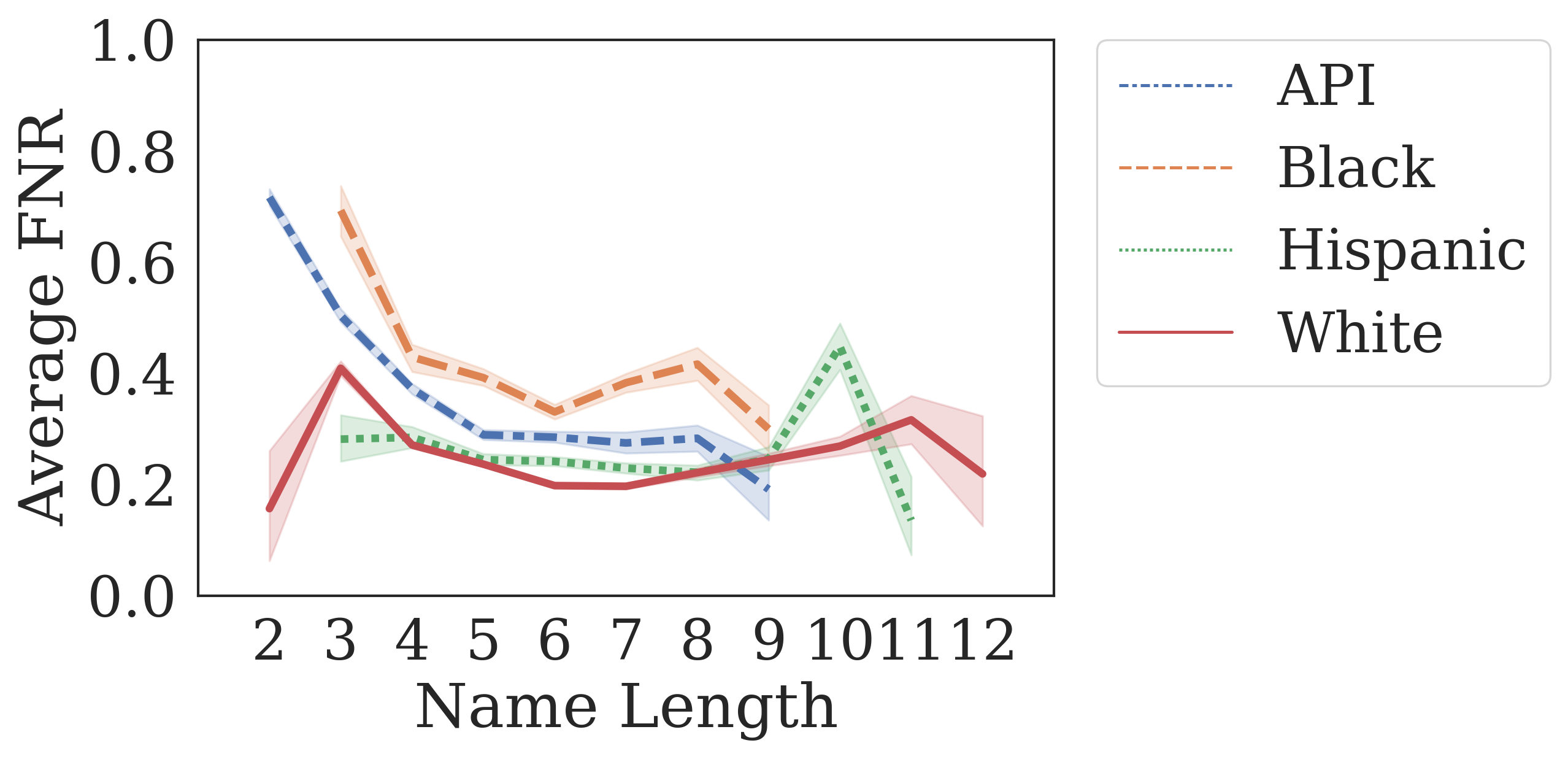}
    \caption{Average FNR across all systems by character length and race/ethnic group.}
    \label{fig:name_len_FNR}
\end{figure}

Table \ref{tab:name-examples} shows examples of names with the highest and lowest FNRs. It is worth noting that API names in LAR2 with high FNR are nearly all 2 characters long. Figure \ref{fig:name_len_FNR} shows the relationship between average FNR across all systems, name length, and group. FNR is lowest for 6-7 character names, and increases as length decreases. However, when matched by character length, API-associated names have higher FNRs than Hispanic and White-associated names nearly across the board. There appear to be higher penalties for short names in the API and Black groups. 


\begin{table*}
\begin{center}
    \begin{tabular}{ |p{1.5cm}|p{6cm}|p{6cm}| } 
         \hline
          & Low FNR & High FNR \\ \hline
         LAR1 & Bob (H), Kristan (W), Vicki (W), Nickie (W), Bethann (W) & German (H), Houston (W), Denver (W), Royal (W), Said (W) \\ \hline
         LAR2 & Maher (W), Nguyen (A), Rajesh (A), Nicoletta (W), Jayesh (A) & Man (A), My (A), In (A), Do (A), So (A)\\ \hline
         NYC & Kaylie (H/F), Keith (W/M), Lena (W/F), Brody (W/M), Brendan (W/F) & Egypt (B/F), Empress (B/F), Asia (B/F), Major (B/M), Malaysia (B/F) \\ \hline
          Congress & Louie Gohmert (W/M), Deborah Ross (W/F), Diana DeGette (W/F), Fred Keller (W/M), Dianne Feinstein (W/F) & Lisa Blunt Rochester (A/F), Aumua Amata Radewagon (A/F), A. Ferguson (W/M), A. McEachin (B/M), Young Kim (A/F) \\
          \hline
    \end{tabular}
\caption{A sample of names with the highest and lowest FNR on average per each dataset.  Race/ethnicity is abbreviated as API (A), Black (B), Hispanic (H), and White (W), while gender is abbreviated female (F), male (M).}
\label{tab:name-examples}
\end{center}
\end{table*}


\begin{figure}
    \includegraphics[width=0.49\textwidth]{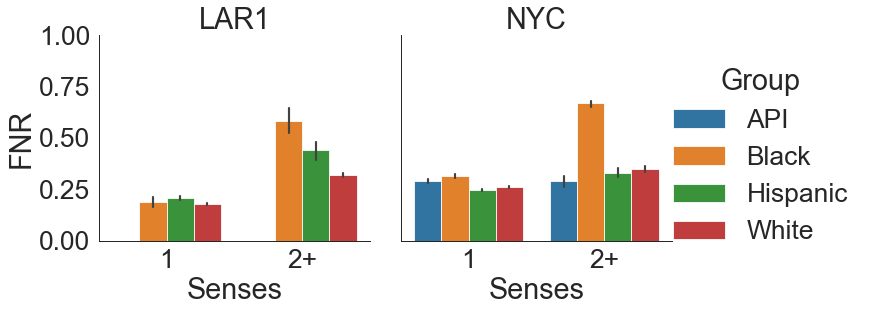}
    \caption{FNR for names with one or multiple word senses (i.e. including non-person word senses)}
    \label{fig:name_ambig}
\end{figure}

High FNR names in Table \ref{tab:name-examples} tend to coincide with other word senses in English. Many are location words (e.g. German, Rochester, Asia). Others double as verbs (`Said'), adjectives (`Young'), nouns (`Major'), and function words (`In'). Using WordNet \citep{wordnet}, a lexical database of English, we examine given names that have overlapping (non-person) senses. Potentially ambiguous given names have a 42\% FNR compared to 24\% for non-ambiguous names. However, the penalty of having an ambiguous name is not the same across groups. Figure \ref{fig:name_ambig} shows that there is a large performance disparity for Black names with multiple senses. This is seen anecdotally in names with similar syntactic/semantic content. For instance, the name `Joy' (API) has a 60\% lower FNR (averaged across systems) than `Blessing' (Black), and `Georgia' (White) has a 25\% lower FNR than `Egypt' (Black).

\section{Discussion}\label{sec:discussion}

This paper considers differences in the performance of three PII maskers on recognizing and redacting names based on demographic characteristics. Supported by quantitative results and error analysis, we find disparities in the fairness of name masking across groups.

In terms of race and ethnicity, API-associated names are often poorly masked. Disparities are shown to be significant for AWS and GCP systems. This is not simply a result of the popularity of the names, as the frequency-controlled LAR1 dataset revealed disparities between API and White-associated names. Name length is considered as a performance factor, but it does not entirely account for the gap between API and White-associated names.


Several systems and datasets show poor performance on the masking of Black-associated names. GCP and Presidio revealed significant differences between Black and White-associated names. Error rates are especially high on the NYC dataset, and are highest for Black women. This is in line with previous research which demonstrates the poor performance of NLP systems on Black women \citep[see inter alia][]{buolamwini2018gender}.



Race and ethnicity were the strongest factors related to PII masking performance, but gender-based differences were also noted. Names which were not strongly associated with gender had the highest error rates. This underscores the importance of considering categories outside the traditional gender binary when evaluating systems for bias.

Of all PII masking systems, the Presidio model (with roBERTa NER) shows fewer significant discrepancies based on demographics. However, all systems demonstrate some significant disparities. Across datasets, the performance difference between groups is not consistent. For instance, the AWS model has poor performance on API names in the LAR2 dataset but not in NYC. We consider this not an issue, but a feature of our evaluation across datasets. The datasets we've chosen contain variety in age groups, locations, and contexts. We argue that evaluating NLP systems responsibly requires careful curation of data, including steps to consider the context of the system and the diverse set of system users and stakeholders.

The aggregate name data used here is openly available and can be used for testing on PII masking, NER, and related systems. We are releasing our templates and code used for sampling data. However, we strongly condemn the use of these datasets for predictive purposes, such as identifying a person's race/ethnicity or gender on the basis of their name without their consent. While our collection of name data forms one of the most comprehensive sets of aggregate names and demographic information available, we are limited by availability of data. The sample of Indigenous and mixed-race names was small, and names were sampled almost exclusively from US-born citizens. In the future, we would like to consider collaborating with the public by developing a database where individuals may actively choose to contribute their name and self-identified information for research.

\section{Conclusion}\label{sec:conclusion}

This work considers the performance of PII masking systems on names sourced from real data. We find disparities related to demographic characteristics, especially race and ethnicity, across all systems. While features such as name length and ambiguity play a role in recognition, they do not fully account for performance differences. Disparities in the performance of PII masking systems reflect historical inequities in the ``Right to Privacy''. The NLP community, as a commodifier of both models and data, has a responsibility to develop more equitable systems to protect the data privacy of all individuals.

\section*{Acknowledgments}\label{sec:acks}
The authors thank Emily M. Bender, Joe Bradley, Chris Brew, Andrew Maurer, and the anonymous reviewers for their helpful comments. 

At different points over the course of the work presented in this paper, A.P. was supported by a research internship at LivePerson, Inc. and also by the National Institutes of Health, National Library of Medicine (NLM) Biomedical and Health Informatics Training Program at the University of Washington (Grant Nr. T15LM007442). The content is solely the responsibility of the authors and does not necessarily represent the official views of the National Institutes of Health.

\clearpage



\clearpage

\appendix

\section{Appendices}
\label{sec:appendix}

\subsection{Post-hoc testing} \label{sec:posthoc}

\smallskip

\begin{strip}
Nemenyi post-hoc significance testing for each dataset.  Significance for each respective system is marked with their respective abbreviation: AWS Comprehend (A), GCP DLP (G), and Microsoft Presidio (P). A `-' indicates a \textit{p}-value above the significance threshold
\end{strip}

\begin{table}[htbp]
    \begin{tabular}{ |l|c|c|c| } 
         \hline
          & Black & Hispanic & White \\ \hline
          Black & - - - & AG - & - GP \\
          Hispanic & AG - & - - - & AG - \\ 
          White & - GP & AG - & - - - \\ 
         \hline
    \end{tabular}
\caption{LAR1 dataset with race/ethnicity}
\end{table}

\begin{table}[htbp]
    \begin{tabular}{ |l c|l l|l l|l l|l l| } 
         \hline
          & & \multicolumn{2}{c|}{API} & \multicolumn{2}{c|}{Black} & \multicolumn{2}{c|}{Hispanic} & \multicolumn{2}{c|}{White} \\ 
          & & F & M & F & M & F & M & F & M \\ \hline
          API & F & - - - & - - - & AG - & A - - & - - - & - G - & A - - & AG - \\
           & M & - - - & - - - & A - - & A - - & - - - & - G - & - G - & AG - \\ \hline
          Black & F & AG - & A - -  & - - -  & - - -  & AG - & AG - & - G - & - G - \\
           & M & A - - & A - - & - - - & - - - & AG - & AG - & - G - & - G - \\ \hline
          Hispanic & F & - - - & - - - & AG - & AG - & - - - & - - - & A - - & AG - \\
           & M & - G - & - G - & AG - & AG - & - - - & - - - & - - - & A - - \\ \hline 
          White & F & A - - & - G - & - G - & - G - & A - - & - - - & - - - & - - - \\
           & M & AG - & AG - & - G - & - G - & AG - & A - - & - - - & - - - \\
         \hline
    \end{tabular}
\caption{NYC dataset with gender, race/ethnicitiy}
\end{table}

\begin{table}[htbp]
    \begin{tabular}{ |l|c|c|c|c| } 
         \hline
          & API & Black & Hispanic & White \\ \hline
          API & - - - & A - - & AG - & AG - \\
          Black & A - - & - - - & - - - & - G - \\
          Hispanic & AG - & - - - & - - - & - - - \\ 
          White & AG - & - G - & - - - & - - - \\
         \hline
    \end{tabular}
\caption{Congress dataset with race/ethnicity.  The Presidio model did not differ significantly based on race/ethnic group.}
\end{table}

\clearpage

\subsection{Frequency sampling}\label{sec:appendix_freq}


This appendix describes in more detail the frequency matching between race/ethnicity groups in the LAR dataset.  The mean observation frequencies for each group are shown in Table \ref{tab:LAR_commonness}. Because there are initially fewer Black-associated names (n=21), we sample all groups to target this smaller distribution. By filtering with a minimum observation size of 2K and maximum observation size of 150K, we achieve similar distributions across groups. However, API names are too sparse under these conditions to be included, and we choose to resample them separately. A Mann-Whitney U test does not find significant differences in frequency between Black, Hispanic, and White-associated names under these conditions (with a threshold of $p=0.05$). A plot of the distributions of this set, which we refer to as LAR1, is shown in Figure \ref{fig:norm_LAR1}.

For API names, we generate a second name set, which we refer to as LAR2. We sample from other groups, using an exponential distribution ($\lambda=480$) that best approximates the API distribution. Only White-associated names maintain $>$20 names under these sampling conditions. A Mann-Whitney U test does not find significant differences between frequencies of API and White groups. Distributions of this set are shown in Figure \ref{fig:norm_LAR2}. 

\medskip


\begin{table}[hb]
\begin{center}
    \begin{tabular}{ |l|r| } 
         \hline
         Group & N \\ \hline
         API & 488 \\ 
         Black & 21573 \\ 
         Hispanic & 25122 \\
         White & 41060 \\ 
         \hline
    \end{tabular}
\caption{Average observation size per name for each race/ethnicity group in the LAR dataset without resampling.}
\label{tab:LAR_commonness}
\end{center}
\end{table}

\newsavebox{\tempbox}
\begin{figure}
\centering
\sbox{\tempbox}{\includegraphics[width=0.48\textwidth]{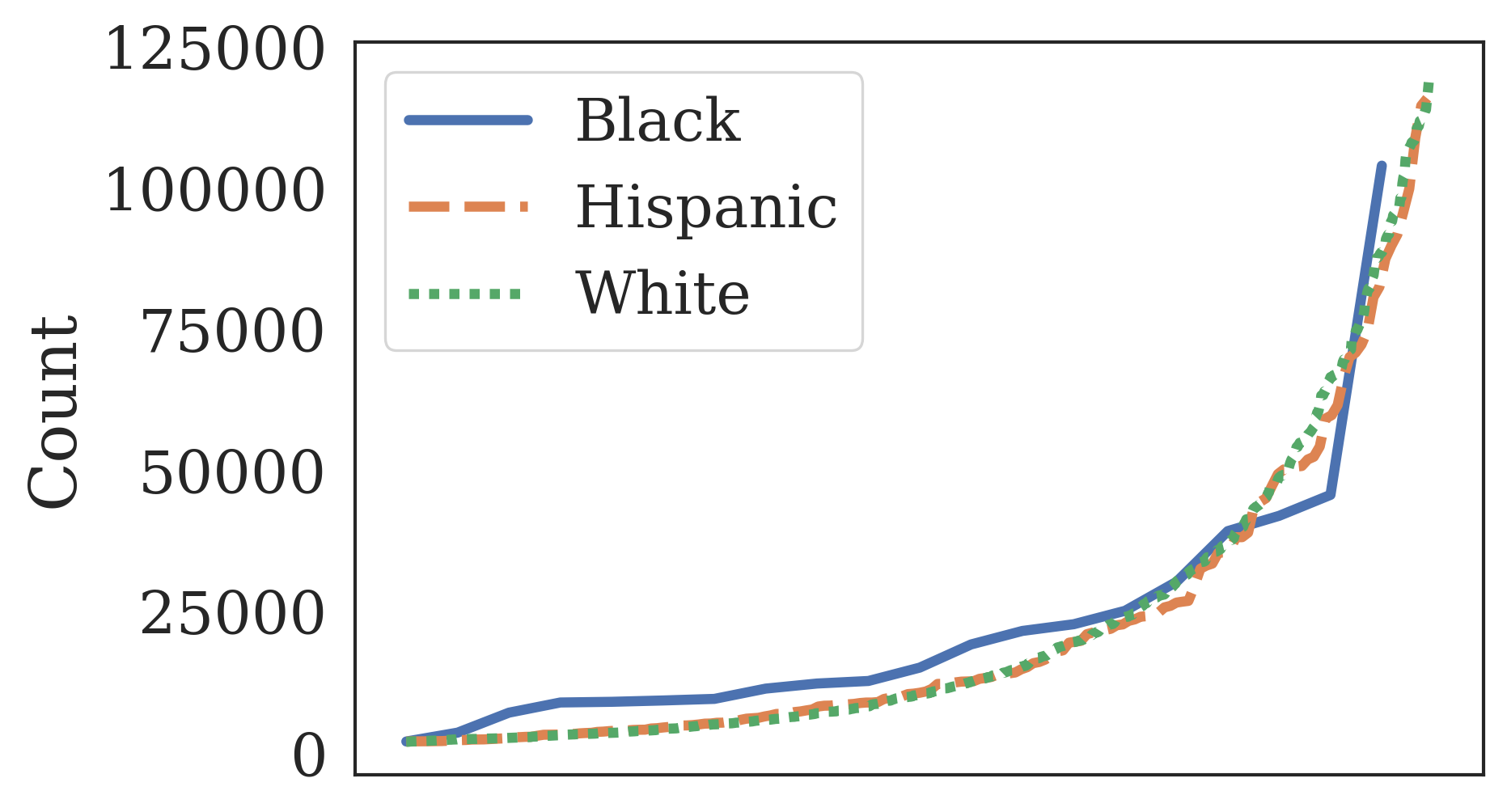}}
\subfloat[Black, Hispanic, and White race/ethnicity groups in LAR1]{\usebox{\tempbox}\label{fig:norm_LAR1}}%
\qquad
\subfloat[API and White race/ethnicity groups in LAR2]{\vbox to \ht\tempbox{%
  \vfil
  \includegraphics[width=0.45\textwidth]{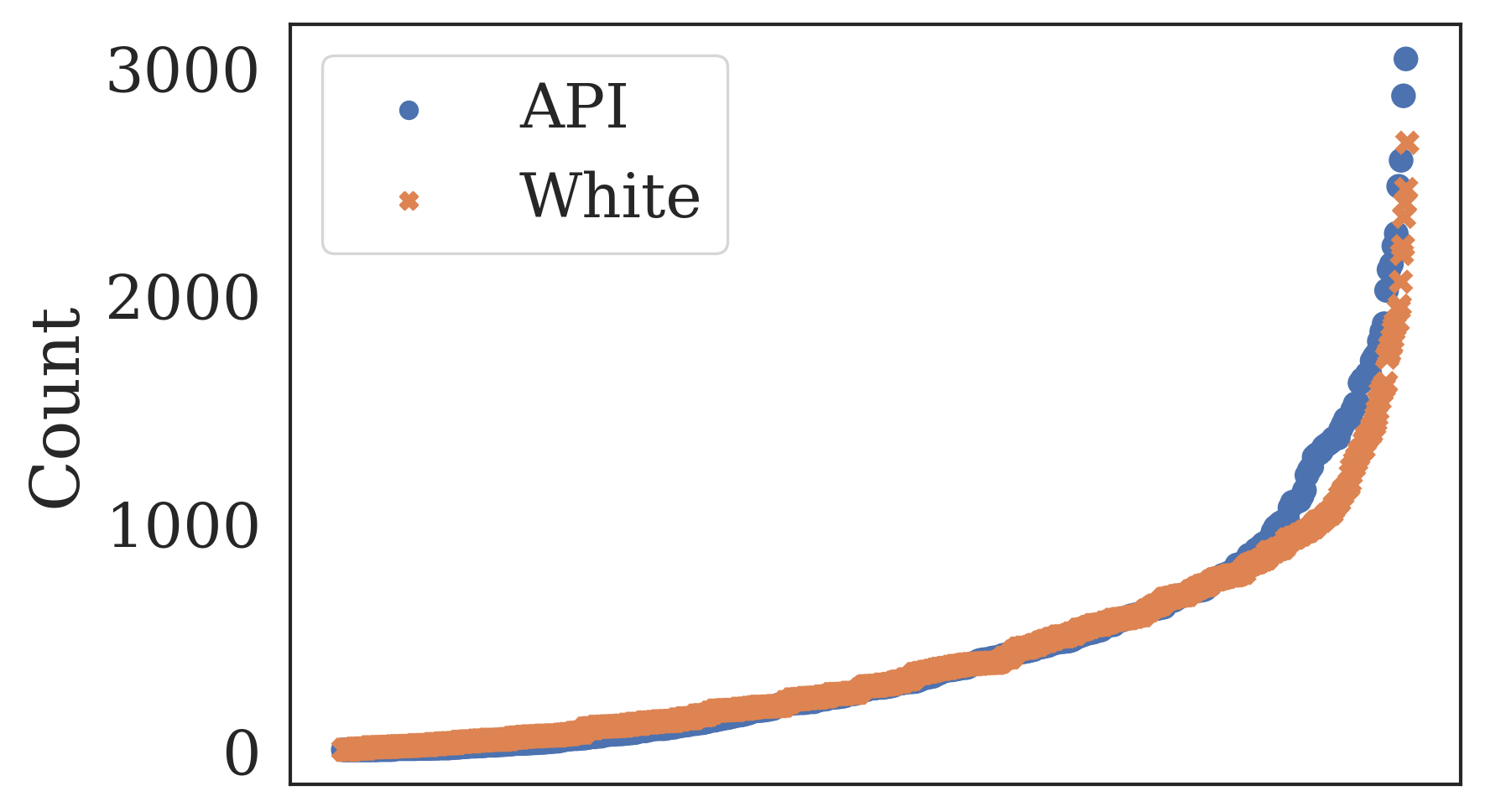}
  \vfil}\label{fig:norm_LAR2}}%
  \caption{Plots of frequency distributions for frequency-matched names from LAR.}
\end{figure}

\clearpage

\subsection{Templates}
\label{sec:appendix_templates}

\begin{table}[ht]
\begin{tabular}{|p{0.8cm}|p{14cm}|}
\hline
\# & Template \\ \hline
1  & Name: \{\{Name\}\} Vouchers:10000200007400001 10000200005000001 \\ \hline
2  & sysmsg1\_\{\{Name\}\}\_ has joined the conversation, \\ \hline
3  & Craig G: 1F to LAS and 2F to SAN \{\{Name\}\} 1D to LAS and 2D to SAN \\ \hline
4  & \{\{Name\}\}    03 caramel beige    is my another foundation \\ \hline
5  & i put in an order on line for \{\{Name\}\} original large size and a code for 20 present off of the 117.00 but it would not take \\ \hline
6  & Hi \{\{Name\}\}! Can you help me with my above question? \\ \hline
7  & hi im \{\{Name\}\} \\ \hline
8  & \{\{Name\}\} isle Jake window \\ \hline
9  & Virtual Assistant : Hi \{\{Name\}\}, how can I help you today? \\ \hline
10 & Thank you, \{\{Name\}\} \\ \hline
11 & this was from \{\{Name\}\} \\ \hline
12 & I think it’s \{\{Name\}\} \\ \hline
13 & Ok, will we receive \{\{Name\}\}’s by that date and at that address as well? \\ \hline
14 & \{\{Name\}\}. Very upset at the moment. I placed two request online to have this order cancelled and I just refused an item from FedEX from your store. \\ \hline
15 & Hello \{\{Name\}\}, Im just trying to get some info on the item I ordered \\ \hline
16 & \{\{Name\}\} (I) paid for the ticket \\ \hline
17 & sysmsg2\_\{\{Name\}\}\_ has left the conversation \\ \hline
18 & hey I lost connection from my previous chat with \{\{Name\}\} \\ \hline
19 & Virtual Assistant : Hi \{\{Name\}\}, we’ll use automated messages to chat with you and Customer Care Professionals are standing by. In a short sentence, let me know how I can help you today \\ \hline
20 & thank you very much \{\{Name\}\}. nice chatting with you! \\ \hline
21 & well .. thank u so much \{\{Name\}\} .. \\ \hline
22 & Did \{\{Name\}\} catch you up on everything? \\ \hline
23 & I was working with \{\{Name\}\} earlier on this chat \\ \hline
24 & The response is signed \{\{Name\}\} \\ \hline
25 & it's YGDFEA the reservation. \{\{Name\}\} \\ \hline
26 & My name is \{\{Name\}\}. I messaged yesterday and have not received a response from anyone \\ \hline
27 & \{\{Name\}\} and I divorced. \\ \hline
28 & do you care that something holy to me was in my food \{\{Name\}\}? \\ \hline
29 & \{\{Name\}\} was very kind and helpful! \\ \hline
30 & oh no \{\{Name\}\} sorry to confuse you \\ \hline
31 & the order is under \{\{Name\}\} \\ \hline
32 & \{\{Name\}\}, one question, when i logged into the App, it shows balance as \$50.. is it USD or CAD?\\ \hline
\end{tabular}
\end{table}

\end{document}